%% file: main.tex
\documentclass[11pt]{article}

\usepackage[final]{acl}

\usepackage{times}
\usepackage{latexsym}
\usepackage[utf8]{inputenc}
\usepackage[T1]{fontenc}
\usepackage{microtype}
\usepackage{ragged2e}
\usepackage{inconsolata}

\usepackage{hyperref}
\usepackage{url}
\usepackage{booktabs}
\usepackage{amsfonts}
\usepackage{amsmath}
\usepackage{nicefrac}
\usepackage{xcolor}
\usepackage{graphicx}
\usepackage{subcaption}
\usepackage{array}

\graphicspath{{figures/}}

\title{Composition Collapse: Stable Factual Knowledge \\
  Does Not Imply Compositional Reasoning}

\author{
  \textbf{Zhe Yu}$^{2*}$ \quad 
  \textbf{Wenpeng Xing}$^{1,2*}$ \quad 
  \textbf{Yunzhao Wei}$^{2}$ \quad 
  \textbf{Jie Chen}$^{3}$ \quad \\
  \textbf{Hongzhi Wang}$^{4}$ \quad 
  \textbf{Xuyang Teng}$^{5}$ \quad 
  \textbf{Meng Han}$^{1,2,6}$ \\
  \rule{0pt}{2.5ex} 
  $^{1}$Zhejiang University \quad $^{2}$Binjiang Institute of Zhejiang University \\ $^{3}$Hong Kong Baptist University \quad $^{4}$Harbin Institute of Technology \quad $^{5}$Hangzhou Dianzi University \\ $^{6}$GenTel.io \\
  \rule{0pt}{1.8ex} 
  \small $^{*}$Equal contribution 
  \vspace{-2.5ex} 
}

\begin{document}

\maketitle

\input{sections/abstract}
\input{sections/introduction}
\input{sections/related_work}
\input{sections/method}
\input{sections/experiments}

\input{sections/causal_grpo}
\input{sections/mechanism}
\input{sections/discussion}

\input{sections/conclusion}

\input{sections/limitations}

\bibliography{references}

\appendix
\input{sections/appendix}

\end{document}

%% file: sections/abstract.tex
\begin{abstract}
Post-training is routinely evaluated through aggregate benchmark
scores that treat multi-hop reasoning as a single capability---as if
a model that answers more questions correctly must be better at
assembling facts.  We show that this assumption can be misleading:
recipes with statistically indistinguishable atomic knowledge produce
composition behaviour separated by over 40 percentage points, a
phenomenon we call \emph{composition collapse}: the systematic failure
to assemble stably-known facts into chains, invisible to aggregate
metrics.  We introduce a double-gate protocol that changes the
estimand from an aggregate compositionality gap to residual
composition failure conditioned on stable atomic access, decomposing
post-training gains into three independent channels: atomic stability,
residual composition, and critical depth.  On a benchmark of temporal
factual chains spanning depths 2--11 across four post-training
recipes, this decomposition reveals that post-training objectives
shift composition capability in directions that aggregate metrics
mask, and suggests that claims about multi-hop reasoning improvement
should be accompanied by atomic-gate-controlled composition metrics.
Diagnostic probes further show that a substantial share of measured
composition failure reflects generation-time computation constraints
rather than permanent inability to compose.
\end{abstract}

%% file: sections/introduction.tex
\section{Introduction}

When a language model fails a multi-hop question---\emph{which came
first, the telephone or flight?}---two explanations are possible.
Either the model does not reliably know the dates, or it knows both
and still cannot compare them.  These demand different fixes: more
factual knowledge versus better compositional reasoning.  Yet
standard evaluation practice treats them as the same thing.

The field evaluates multi-hop reasoning through aggregate benchmark
accuracy on datasets like HotpotQA~\citep{yang2018hotpotqa}: a model that answers more composite questions correctly is
assumed to be better at composing facts.  The
\emph{compositionality gap} \citep{press2023measuring,
dziri2023faith}---the drop from single-hop to multi-hop
accuracy---is widely used to quantify composition failure.  But
this metric conflates two categorically different failure modes.  A
model may answer a sub-question correctly once without \emph{stable}
access to the underlying fact, yet a single correct answer does not
guarantee the model can reproduce that knowledge in a longer chain.  Attributing the entire gap to
``composition failure'' mixes atomic-instability variance into what
is claimed as a reasoning deficit.  Prior work has shown that
knowledge extraction can fail systematically even when facts are
present in the training distribution \citep{allenzhu2023physics},
and that multi-hop failures can be localised to specific layer-level
timing problems \citep{biran2024hopping}.  Recent work on shortcut-free
latent multi-hop evaluation (SOCRATES; \citet{yang2025socrates}) shows that
composition success varies dramatically by entity type even when atomic
facts are known---reinforcing the need to measure composition net of atom
confidence.  These studies establish
that reasoning failures exist independently of memorisation---a
premise we share.  But they do not provide a measurement protocol
that cleanly separates the two failure modes at evaluation time.

This conflation is not a minor measurement issue.  Two post-training
recipes whose atomic stability differs by less than two percentage
points can diverge by over 40~pp in residual composition failure at
the shallowest non-trivial depth.  We call this
\emph{composition collapse}: the systematic failure to assemble stably-known
facts into correct chains, hidden from aggregate metrics and visible
only when atomic knowledge is controlled for.

We introduce a \emph{double-gate protocol} that changes the estimand
from an aggregate gap to residual composition failure conditioned on
stable atomic access.  Before measuring whether a model can compose
facts, the protocol first verifies that it stably possesses each fact
(consistency across paraphrases) and can answer each sub-question in
isolation.  Errors that survive both gates are \emph{residual
composition failure}: the model stably knows the parts, has verbalised
them one at a time, and still fails to assemble them.  With this
protocol, post-training gains decompose into three independent
channels: $\Delta_{\text{atom}}$ (atomic stability),
$\Delta_{\text{comp}}$ (residual composition at matched atoms), and
$\Delta_{\text{depth}}$ (the depth at which residual failure exceeds
50\%).  We instantiate the protocol on \textsc{D4v2}, a 390-question
temporal benchmark spanning four task families at depths 2--11.

The benchmark covers temporal factual chains (dates, events, causal
order); domain generality beyond temporal reasoning remains open and
is examined preliminarily through in-context probes and a cross-domain
pilot (\S\ref{sec:experiments}).

We evaluate six open-weights models at 7--13B scale spanning four
post-training recipes---base, RLHF~\citep{ouyang2022training}, SFT reasoning-trace distillation,
and native outcome-verified RL---and run a controlled same-base \citet{lora} LoRA
intervention (SFT-answer, SFT-trace, \citet{deepseekmath} GRPO) that isolates the training
objective from base-model confounds.  Three findings emerge.  First,
base models are filtered by atomic knowledge, not
composition---their atomic gate admits ${<}5$ cases at any depth.  Second, at matched
atomic stability, residual failure ranges from 0\% to 47\% at depth~2
depending on the training objective, and outcome-verified RL
outperforms SFT distillation by a wide margin under identical data and
compute.  Third, training on a composition depth closes most of that
depth's residual failure gap but transfers only weakly to adjacent
held-out depths.  Enabling chain-of-thought reasoning reduces residual
failure by 3--5$\times$ (\S\ref{sec:additional}), locating much of the
bottleneck in generation-time computation, not in static knowledge
representation.  Our contributions are: (i)~a double-gate protocol
that provides a cleaner measurement of composition net of atomic
knowledge; (ii)~a three-channel decomposition that makes post-training
recipes comparable on a like-for-like basis; and (iii)~a controlled
causal intervention isolating a post-training-specific composition
channel with limited depth transfer.  We do not claim to reveal the
essence of post-training; rather, we offer a measurement protocol that
exposes previously hidden variation across recipes at matched atoms.

%% file: sections/related_work.tex
\section{Related Work}

\textbf{Measuring the compositionality gap.}
\citet{press2023measuring} formalised the compositionality gap as the
drop from single-hop to multi-hop accuracy; \citet{dziri2023faith}
showed Transformer performance on multi-hop composition degrades with
chain length.  \citet{yang2025socrates} introduced latent multi-hop
probes (SOCRATES) that bypass surface shortcuts, showing composition
success varies dramatically by entity type even when atomic facts are
known.  These frameworks share a confound: the single-hop baseline
does not verify that facts are \emph{stably} accessible---a single
correct sub-question answer does not guarantee the model can reproduce
that knowledge when the same fact appears embedded in a chain.  Our
double-gate protocol adds a per-fact stability filter that separates
transient retrieval from stable knowledge, converting a
population-level correction into a per-case admissibility test.

\begin{figure*}[t]
  \centering
  \includegraphics[width=0.95\linewidth]{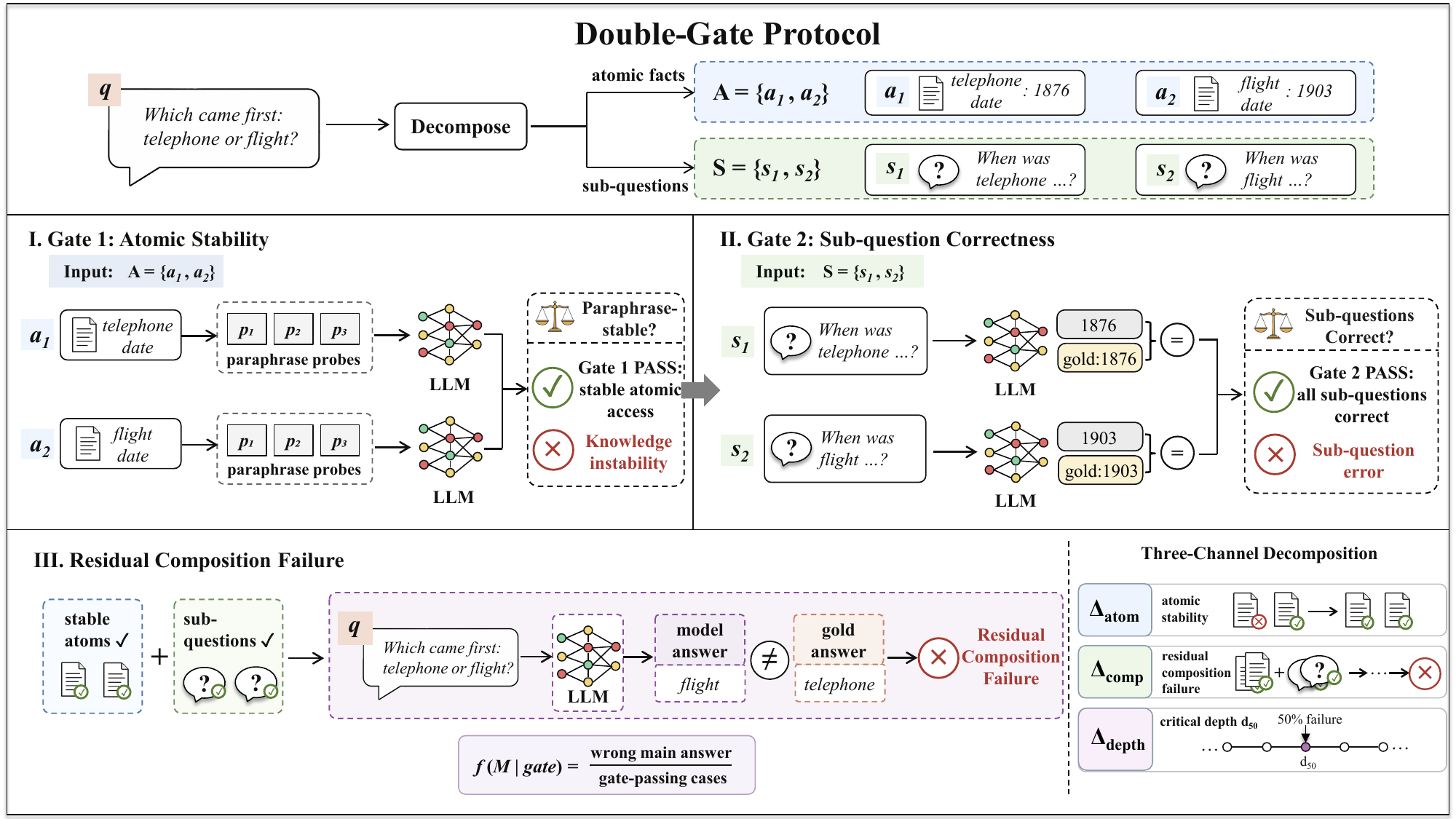}
  \caption{\textbf{Double-gate protocol for measuring residual
  composition failure.}  A multi-hop question is decomposed into
  atomic facts and aligned sub-questions.  \textbf{I.} Atomic
  Stability verifies stable and correct access to each atom across
  paraphrases.  \textbf{II.} Sub-question Correctness verifies that
  each sub-question is answered correctly in isolation.
  \textbf{III.} Residual Composition Failure measures whether the
  original multi-hop question is still answered incorrectly after
  both gates are passed.  This separates composition errors from
  unstable knowledge access and supports the three-channel
  decomposition into $\Delta_{\text{atom}}$,
  $\Delta_{\text{comp}}$, and $\Delta_{\text{depth}}$.}
  \label{fig:framework}
\end{figure*}

\textbf{Post-training and reasoning.}
RLHF~\citep{ouyang2022training}, SFT reasoning-trace
distillation~\citep{wei2022chain,deepseek_r1}, and
outcome-verified RL~\citep{deepseekmath} are the dominant paradigms for
improving reasoning.  Prior work evaluates these recipes through
aggregate benchmark scores, which conflate knowledge acquisition with
reasoning improvement.  Our three-channel decomposition shows these
recipes shift atomic stability and residual composition independently:
atomic stability saturates after the base-to-instruct jump, while
residual composition varies by 40+~pp across recipes at matched
atoms.  This decomposition is orthogonal to the training objective: it
provides a measurement lens, not a modelling claim.

\textbf{Computation and composition.}
\citet{biran2024hopping} localised multi-hop failure to layer-level
timing---the model encodes both facts but processes the second too
late.  \citet{allenzhu2023physics} showed knowledge extraction fails
systematically even when facts are in the training distribution, and
CoT data during training can bridge the gap.  Our CoT diagnostic
(\S\ref{sec:additional}) complements these findings: enabling CoT at
inference time recovers 70--75\% of gate-passing failures, locating
much of the residual failure in generation-time computation rather
than permanent representational deficit.  Together with the patching
experiment (Appendix~\ref{app:mechanism}), which finds negligible
information in prompt-end representations, these results triangulate
the bottleneck to the generation process itself.

%% file: sections/method.tex
\section{Method}
\label{sec:method}

Figure~\ref{fig:framework} summarises the pipeline.  A composition
instance is a main question $q$ whose answer requires $k$ atomic facts
$a_1,\ldots,a_k$ (the depth).  Each atom carries $m_i$ paraphrased
probes plus a sub-question $s_i$ matching how the atom appears in the
chain.  An instance \emph{passes the atomic gate} iff every atom is
answered consistently across all paraphrases (consistency matcher of
\citealp{press2023measuring}); it \emph{passes the sub-question gate}
iff every $s_i$ is correct in isolation.  Our primary quantity is the
\emph{residual composition failure rate}: the fraction of gate-passing
instances whose main answer is still wrong.  By construction it
upper-bounds pure composition failure---format artifacts, distractor
attention, and greedy-trajectory degeneracy can inflate the measured
rate without reflecting genuine compositional inability (we quantify
this margin in \S\ref{sec:discussion}).

Given a recipe $R$ and a matched comparator $R'$ (atomic stabilities
within 2~pp), we decompose the post-training benefit into three
channels:
\begin{align*}
\Delta_{\text{atom}}(R) &= s(M_R) - s(M_{R'}), \\
\Delta_{\text{comp}}(R) &= f(M_R \mid \text{gate}) - f(M_{R'} \mid \text{gate}), \\
\Delta_{\text{depth}}(R) &= d_{50}(M_R) - d_{50}(M_{R'}),
\end{align*}
where $s$ is atom stability, $f$ is residual composition failure, and
$d_{50}$ is the depth at which $f$ first exceeds 50\%.  When atomic
stabilities are matched, $\Delta_{\text{comp}}$ isolates a pure
composition benefit; the $\Delta_{\text{comp}}$--$\Delta_{\text{depth}}$
gap later lets us distinguish a training-data effect from a
model-structure effect.

\paragraph{Benchmark.}
\textsc{D4v2} contains 390 main questions and 2\,490 atomic
sub-questions built from four task families: a depth-2 pair-ordering
control and three depth-covering families
(\textsc{temporal\_rank}, \textsc{temporal\_successor},
\textsc{temporal\_interval\_decoy}) at $d{\in}\{4,6,8\}$, with
\textsc{temporal\_interval\_decoy} additionally sampling intermediate
fact counts ($d{=}7,9,11$; $n{=}30$ each).  Task families differ in
required chain structure---ranking a set, stepping forward in time,
and disambiguating overlapping intervals---so same-depth cases are not
interchangeable in difficulty.  To enable cross-family comparison at
shared depth bins, we group depth-7 cases with depth~4, depth~9 with
depth~6, and depth~11 with depth~8 (see Appendix~\ref{app:d4v2} for
rationale and full breakdown).  Three additional synthetic task
families (\textsc{kinship}, \textsc{numerical}, \textsc{spatial})
built from fictional entities form a separate evaluation suite; as
discussed in \S\ref{sec:experiments}, they are excluded because our
closed-book protocol cannot evaluate composition when internal atomic
knowledge is absent.

\paragraph{Recipe coverage and evaluation.}
We evaluate four post-training recipes: base, RLHF, SFT distillation
of reasoning traces, and native outcome-verified RL (RLVR), across six
open-weights models at the 7--13B scale (checkpoints in
Appendix~\ref{app:atoms}).  All numbers use the Press et al.\
consistency matcher; full adjudication tiers and vLLM vs.\ HuggingFace
cross-validation are in Appendix~\ref{app:e2} and~\ref{app:vllm_hf}.
Inference is greedy ($T{=}0$), XML-structured.

%% file: sections/experiments.tex
\section{Experiments and Results}
\label{sec:experiments}

We report three observations that rule out the most obvious
alternative explanations and isolate a post-training-specific
composition channel.  First, the atomic gate removes base models
entirely, so the failures we measure are not knowledge gaps
(\S\ref{sec:experiments}, paragraph~1).  Second, two recipes with
statistically indistinguishable atomic knowledge produce
categorically different composition at the shallowest non-trivial
depth (paragraph~2).  Third, within a fixed depth, task structure
changes residual failure by up to 66~percentage points, so depth is
not the dominant axis of difficulty (paragraph~3).

\begin{figure}[t]
  \centering
  \includegraphics[width=0.95\linewidth]{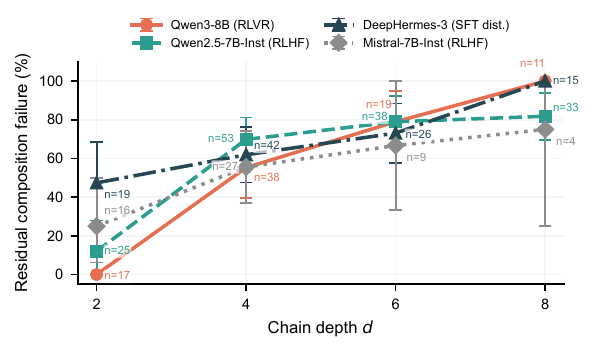}
  \caption{\textbf{The composition collapse curve.}  Residual composition
  failure rate versus depth on \textsc{D4v2} (bootstrap 95\% CI).  All
  gate-passing recipes cross 50\% failure by depth 4--5; two reach 100\%
  by depth~8.  Base models are omitted (see Appendix~\ref{app:atoms});
  their atomic gate admits $<$5 cases at every depth.  CI width grows
  with depth as gate-passing $n$ declines; points at depth~8 should be
  interpreted with particular caution ($n \leq 37$ for all recipes).}
  \label{fig:collapse}
\end{figure}

Figure~\ref{fig:collapse} shows residual composition failure as a
function of chain depth, with bootstrap 95\% CIs over atomic-gate
passing cases.

\begin{table}[t]
\centering\footnotesize
\setlength{\tabcolsep}{3pt}
\caption{Residual composition failure rate (\%) by depth on
\textsc{D4v2}, restricted to gate-passing post-trained models.
``Atom stab.'' is the per-probe atomic-stability rate; the
atomic gate imposes a stricter per-fact criterion, so the effective
fact-level pass rate is lower.  Parenthesised values are $n$,
the number of atomic-gate-passing cases.  We omit $n{<}5$.}
\label{tab:main}
\resizebox{\columnwidth}{!}{%
\begin{tabular}{lcrrrr}
\toprule
Model & Stab. (\%) & $d=2$ & $d=4$ & $d=6$ & $d=8$ \\
\midrule
Qwen3-8B \citep{qwen3} (RLVR)                & 87.7 & \phantom{0}0.0~(17) & 55.3~(38) & 78.9~(19) & 100.0~(11)$^{\dagger}$ \\
Qwen2.5-7B-Inst \citep{qwen25} (RLHF)     & 89.9 & 12.0~(25)           & 69.8~(53) & 78.9~(38) & \phantom{0}81.1~(37) \\
DeepHermes-3 \citep{deephermes} (SFT distill) & 88.0 & 47.4~(19)           & 61.9~(42) & 73.1~(26) & 100.0~(15)$^{\dagger}$ \\
Mistral-7B-Inst \citep{mistral7b} (RLHF)     & 78.8 & 25.0~(16)           & 55.6~(27) & 66.7~(\phantom{0}9)$^{\dagger}$  & -- \\
\bottomrule
\end{tabular}%
}
{\footnotesize\vspace{4pt} $^{\dagger}$\, $n < 15$ gate-passing cases; point estimates in these cells should be interpreted with caution---bootstrap CIs exceed $\pm$25~pp.}
\end{table}

\paragraph{Base models are filtered by knowledge, not composition.}
All three pre-RLHF base models we test (Llama-3-8B \citep{llama3}, Mistral-7B-v0.1 \citep{mistral7b},
Qwen2.5-7B \citep{qwen25}) have atomic stability below 15\%, so their joint
stability at depth $k$ is near zero and the atomic gate admits fewer
than five cases at any depth.  Attributing their failure to
``composition'' would be a category error---precisely the
confound our protocol is designed to remove.  Full per-model atomic
rates are reported in Appendix~\ref{app:atoms}.

\paragraph{Synthetic task families are excluded by design.}
Three synthetic families (\textsc{kinship}, \textsc{numerical},
\textsc{spatial}) built from fictional entities are part of the full
suite but are excluded because no model can possess internal knowledge
of entities absent from its training data (atomic accuracy $=$ 0\%);
the protocol correctly prevents composition measurement when atoms are
absent.  We therefore restrict analysis to temporal facts where
models demonstrably possess the requisite knowledge.

\subsubsection{Composition collapse persists under in-context evidence}
To verify that composition difficulty is not merely an artefact of
closed-book knowledge access, we evaluated Qwen2.5-7B-Instruct on the
same synthetic families with all facts supplied in the prompt as
\texttt{[Evidence]} blocks.  Under this in-context protocol, only
21.8\% of cases survive the sub-question gate, and on those
gate-passing cases residual composition failure is 88.9\%
(Table~\ref{tab:synthetic_incontext}).  This convergence---88.9\% in-context
versus 55--100\% closed-book at matched depths---rules out knowledge
retrieval as the dominant explanation for composition collapse.
When the relevant facts are supplied in the prompt and gate-passing cases
are conditioned on correct sub-question use, composition remains the
primary failure mode; the bottleneck is assembly, not lookup.

\begin{table}[t]
\centering\footnotesize
\setlength{\tabcolsep}{4.5pt}
\caption{In-context composition on synthetic families (Qwen2.5-7B-Instruct).
Facts are supplied in the prompt.  ``Atom pass'' = all sub-questions correct.
Residual failure = main-question error given atom pass.}
\label{tab:synthetic_incontext}
\resizebox{\columnwidth}{!}{%
\begin{tabular}{lccc}
\toprule
Task family & $n$ & Atom pass & Residual failure (\%) \\
\midrule
Kinship (matrilineal) & 33 & 8 & 62.5 \\
Kinship (patrilineal) & 33 & 9 & 66.7 \\
Spatial (eastward) & 33 & 12 & 91.7 \\
Spatial (northward) & 33 & 6 & 83.3 \\
Numerical (offset) & 33 & 1 & 100.0 \\
\midrule
Aggregate & 165 & 36 & 88.9 \\
\bottomrule
\end{tabular}%
}
\end{table}

\paragraph{Quantifying the conflation.}
We recompute residual failure using only the sub-question gate (the
standard single-gate protocol).  The aggregate single-gate rate exceeds
the double-gate rate by 2.5--9.8~pp across models, but the cell-level
inflation peaks at 11.4~pp, and the atomic gate removes 15--44\% of
single-gate-passing cases---cases prior work would count as valid
composition trials whose constituent atoms are not stably known.
These findings motivate the double-gate protocol not as a refinement
but as a correction of the estimand.

\paragraph{Matched atoms, divergent composition: the composition collapse.}
Three recipes admitted by the gate cluster tightly on atomic
stability---Qwen3 87.7\%, DeepHermes 88.0\%, Qwen2.5-Instruct
89.9\%---yet collapse to 0\%, 47\%, and 12\% residual composition
failure at depth~2 (Table~\ref{tab:main}).  This
\emph{composition collapse}---a 47-point gap between a native
outcome-verified RL model and a recipe that merely imitates its
reasoning traces---is the paper's cleanest correlational signal: at
the knowledge level these recipes are indistinguishable, at the
composition level they are categorically different.

\paragraph{Limited depth transfer under current recipes.}
Table~\ref{tab:main} shows that all four gate-passing
recipes sit at or above 67\% residual failure by depth~6, and every
reportable measurement at depth~8 is 75\% or higher; Qwen3 and
DeepHermes both reach 100\%.  The best recipe we evaluate pushes
$d_{50}$ no higher than 5--6 hops.  Within our recipe coverage---two
RLHF variants, an SFT reasoning-trace distillation, and a native
RLVR model---the depth ceiling does not meaningfully move, suggesting
that current post-training recipes improve shallow composition more
reliably than depth extrapolation.

\paragraph{Structure dominates depth.}
At fixed depth~4, Qwen3's residual failure ranges from 38.5\% on
\textsc{temporal\_successor} to 92.9\% on
\textsc{temporal\_rank}---a 54-point gap at identical chain length
(Table~\ref{tab:task_depth_main}).  The within-depth spread even
exceeds the within-task depth progression on
\textsc{temporal\_interval\_decoy} (41\% at $d{=}4$ $\to$ 57\% at
$d{=}6$ $\to$ 80\% at $d{=}8$).  The ordering of models by residual
failure reverses across task families, which is incompatible with any
account that treats depth as a monotone difficulty axis.

\begin{table}[t]
\centering\footnotesize
\setlength{\tabcolsep}{4.5pt}
\caption{Residual failure (\%) by task family and depth for Qwen3-8B
(double gate).  At fixed depth, task structure changes failure by up to
54~pp.  Parenthesised values are $n$.  ``--'' = $n{<}5$.}
\label{tab:task_depth_main}
\resizebox{\columnwidth}{!}{%
\begin{tabular}{@{}lcccc@{}}
\toprule
Task family & $d{=}2$ & $d{=}4$ & $d{=}6$ & $d{=}8$ \\
\midrule
Pair control & \phantom{0}0.0~(17) & -- & -- & -- \\
Temp.\ successor & -- & 38.5~(13) & 88.9~(\phantom{0}9) & -- \\
Temp.\ int.\ decoy & -- & 41.2~(17) & 57.1~(\phantom{0}7) & 80.0~(5) \\
Temp.\ rank & -- & 92.9~(14) & 83.3~(\phantom{0}6) & 100.0~(6) \\
\bottomrule
\end{tabular}%
}
\end{table}

\subsection{Decomposition into three channels}

Having established that composition failure survives the atomic gate
and varies dramatically across recipes, we decompose post-training
benefits into the three channels of \S\ref{sec:method}: atomic
stability ($\Delta_{\text{atom}}$), residual composition accuracy at
matched atoms ($\Delta_{\text{comp}}$), and critical depth
($\Delta_{\text{depth}}$).

\subsubsection{\texorpdfstring{$\Delta_{\text{atom}}$}{Delta atom}: post-training buys atomic stability in large steps}
The base-to-instruct jump is 76--88~pp of atomic stability; once paid,
further improvements are single percentage points.  For a 7--13B model,
``does post-training help composition?'' is first a question about
whether the atoms ever become stable enough to attempt composition.

\subsubsection{\texorpdfstring{$\Delta_{\text{comp}}$}{Delta comp}: at matched atoms, recipe matters by 40+pp}
At matched atomic stability (87--90\%), the three gate-passing recipes
diverge by 47~pp at depth~2 (Table~\ref{tab:main}): taking Qwen3 (RLVR)
as reference, $\Delta_{\text{comp}}$ is $-$47.4~pp (DeepHermes) and
$-$12.0~pp (Qwen2.5-Inst).  Two recipes indistinguishable on
knowledge benchmarks differ by 35~pp in residual composition.
The collapse of SFT-distilled reasoning suggests that distilling
reasoning traces can induce surface pattern-matching that generalises
worse than outcome-verified RL.  Base-model confounds for DeepHermes
(Llama-3 vs.\ Qwen2) are partly addressed via R1-distill comparisons in
Appendix~\ref{app:atoms}.

\subsubsection{\texorpdfstring{$\Delta_{\text{depth}}$}{Delta depth}: shallow depth generalisation}
Across post-trained models, $d_{50}$ averages in $[3,5]$
(Table~\ref{tab:critical_depth}).  Even the strongest recipe does not
exceed $d_{50}{=}5$--6, and \textsc{temporal\_rank} pushes $d_{50}$
below 4 for every model.  Post-training improves shallow composition
more reliably than depth extrapolation.

\begin{table}[t]
  \centering\footnotesize
  \setlength{\tabcolsep}{5pt}
  \caption{Critical depth ($d_{50}$) per task family, consistency gate.
  $d_{50}$ is estimated by linear interpolation between the two depth bins
  ($\{2,4,6,8\}$) bracketing 50\% residual failure.  Ranges reflect the
  uncertainty band from the 95\% Clopper--Pearson CI on the underlying
  binomial rates.  All values are approximate due to the coarse 4-bin
  resolution and small gate-passing sample sizes at deeper depths.}
  \label{tab:critical_depth}
  \resizebox{\columnwidth}{!}{%
  \begin{tabular}{lcccc}
  \toprule
  Model & \textsc{rank} & \textsc{succ.} & \textsc{int\_decoy} & Agg. \\
  \midrule
  Qwen3 (RLVR) & $<$4 & [4,6] & [4,6] & [3,5] \\
  Qwen2.5-Inst (RLHF) & $<$4 & [4,6] & [4,6] & [3,5] \\
  DeepHermes (SFT) & $<$4 & $<$4 & $<$4 & $<3$ \\
  Mistral-Inst (RLHF) & $<$4 & $<$4 & [4,5] & $<3$ \\
  \bottomrule
  \end{tabular}%
  }
\end{table}

%% file: sections/causal_grpo.tex
\section{Causal intervention: training reduces seen-depth collapse with limited transfer}
\label{sec:causal_grpo}

The cross-recipe analysis in \S\ref{sec:experiments} is correlational.
To rule out confounding by base model, we train four controlled
LoRA-GRPO variants on
Qwen2.5-7B-Instruct\citep{qwen25}\footnote{All variants use learning rate $10^{-5}$,
8 generations per prompt, reward = consistency match + XML format +
anti-template-echo, \citet{unsloth} (memory-efficient LoRA~\citep{lora}) with patched TRL.
v1/v2: LoRA rank 32, 4 epochs, $\sim$1\,h. v3/v4: LoRA rank 64, 8 epochs,
$\sim$3.2\,h (v3) / $\sim$5\,h (v4).  Details in Appendix~\ref{app:grpo}.}:
v1 trains on depths $\{2,4\}$, v2 on $\{2,4,6\}$, v3 on $\{2,4,6\}$ with
heavier budget (rank 64, 8 epochs), and v4 extends the heavy-budget setup
to train on $\{2,4,6,8\}$.  v1 holds out depths $\{6,8\}$ at test time; v2 and v3 hold out $\{8\}$;
v4, which trains on all four D4V2 depths $\{2,4,6,8\}$, evaluates
depth~8 in-distribution.  Evaluation uses the same protocol as
\S\ref{sec:experiments}.

\begin{figure}[t]
  \centering
  \includegraphics[width=0.95\linewidth]{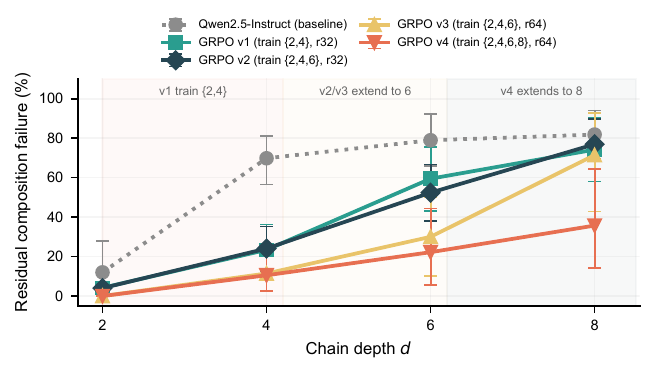}
  \caption{Baseline (Qwen2.5-Instruct, dotted) vs.\ four LoRA-GRPO
  variants.  Shaded bands mark training-depth regions.
  v1/v2 (rank~32, 4~epochs) show that adding depth-6 training
  improves $d{=}6$ but not $d{=}8$; v3 (rank~64, 8~epochs) further
  reduces $d{=}6$ to 27.9\% with minimal $d{=}8$ transfer; v4
  (train $\{2,4,6,8\}$, rank~64, 8~epochs) cuts $d{=}8$ to
  36.0\%, demonstrating that the depth ceiling moves with
  training-data coverage.  All CIs are bootstrap 95\%.}
  \label{fig:grpo_causal}
\end{figure}

\begin{table}[t]
\centering\footnotesize
\setlength{\tabcolsep}{8pt}
\caption{Residual failure (\%) for four GRPO variants vs.\ baseline.
v2 adds depth-6 to training (7~pp improvement at $d{=}6$); v3 uses
heavier budget (rank~64, 8 epochs), cutting $d{=}4$ and $d{=}6$
residual failure by an additional 15--24~pp.  v4 (train $\{2,4,6,8\}$, rank~64, 8 epochs) reduces $d{=}8$ by 45~pp vs.\ baseline and by 36~pp vs.\ v3, showing that training on depth~8 largely closes the depth-8 gap.  Atomic stability remains within 1~pp of
baseline (89.9\%) for all four.}
\label{tab:grpo_v1_v2}
\resizebox{\columnwidth}{!}{%
\begin{tabular}{lcccc}
\toprule
 & $d{=}2$ & $d{=}4$ & $d{=}6$ & $d{=}8$ \\
\midrule
Qwen2.5-Inst baseline & 12.0 & 69.8 & 78.9 & 81.1 \\
GRPO v1 (train $\{2,4\}$, r32) & 4.2 & 24.5 & 59.5 & 74.2 \\
GRPO v2 (train $\{2,4,6\}$, r32) & 4.0 & 24.1 & 52.4 & 76.9 \\
GRPO v3 (train $\{2,4,6\}$, r64) & 0.0 & \textbf{8.5} & 27.9 & 71.9 \\
GRPO v4 (train $\{2,4,6,8\}$, r64) & 0.0 & 7.8 & \textbf{20.6} & \textbf{36.0} \\
\midrule
$\Delta$ v1 & $-$7.8 & $-$45.3 & $-$19.5 & $-$6.9 \\
$\Delta$ v2 & $-$8.0 & $-$45.7 & $-$26.6 & $-$4.2 \\
$\Delta$ v3 & $-$12.0 & $-$61.3 & $-$51.0 & $-$9.2 \\
$\Delta$ v4 & $-$12.0 & $-$62.0 & $-$58.3 & $-$45.1 \\
\bottomrule
\end{tabular}%
}
\end{table}

\paragraph{Four causal conclusions.}
\emph{(1) RL on composition improves composition net of atoms.}
Training cuts depth-4 residual failure by 45--62~pp across variants
while atomic stability moves by 1--3~pp; the effect is a clean
$\Delta_{\text{comp}}$.  At depths 4 and 6, baseline and GRPO v1
bootstrap CIs are disjoint (Figure~\ref{fig:grpo_causal}).
\emph{(2) OOD depth generalisation decays.}  Training on
$\{2,4\}$ generalises 20~pp to depth 6, but only 7~pp to depth 8.
\emph{(3) In-distribution gains do not transfer to deeper hold-out depths---unless the deeper depth is included in training.}
Adding depth-6 to training (v2) improves depth~6 by 7~pp \emph{but does
not shift depth~8 relative to v1}.  Heavier budget without depth-8 data
(v3) cuts $d{=}6$ to 27.9\% but leaves $d{=}8$ at 71.9\%, only 5~pp
below v2.  The ratio of out-of-distribution gain to in-distribution gain
is 0.21, showing that deeper training data alone does not transfer to
farther hold-out depths under a LoRA-GRPO setup.
\emph{(4) Depth-8 failure is trainable when depth-8 data is included.}
v4 (train $\{2,4,6,8\}$, rank~64, 8 epochs) reduces $d{=}8$ residual
failure to 36.0\%---a 45~pp drop from baseline and a 36~pp drop from v3.
All other depths improve modestly (e.g., $d{=}6$ drops from 27.9\% to
20.6\%).  Because v4 evaluates depth~8 in-distribution (depth~8 is in its
training mix), this gain reflects training-coverage extension rather than
generalised depth-8 reasoning capability; it demonstrates that the
ceiling is governed by whether a depth appears in training, not by a
hard architectural limit.  Across all four variants, the clearest signal
is that \emph{training on a depth closes most of that depth's residual
failure gap, but transfers only weakly to adjacent depths}.

\paragraph{Cross-model replication.}
The same GRPO recipe (train $\{2,4,6\}$, LoRA rank~32, 4 epochs)
applied to Llama3-8B-Instruct~\citep{llama3} reduces $d{=}4$ residual failure to
17.9\% and $d{=}6$ to 42.9\%---outperforming the matched Qwen
GRPO~v2 by 13~pp at $d{=}4$ and 15~pp at $d{=}6$
(Table~\ref{tab:cross_model_main}).  The $\Delta_{\text{comp}}$ direction
and magnitude replicate across base-model families, ruling out a
Qwen-specific artifact.  On Mistral-7B-v0.1 \citep{mistral7b} (base), the same GRPO
recipe lifts atomic stability from $<$15\% to a level where 34
cases survive the double gate at $d{=}4$, with 0\% residual
failure---suggesting that GRPO can simultaneously improve atomic
knowledge and composition when starting from a low base, though the
small gate-passing $n$ warrants caution.

\begin{table}[t]
\centering\footnotesize
\setlength{\tabcolsep}{5pt}
\caption{Same GRPO recipe (train $\{2,4,6\}$, rank~32, 4 epochs) on two
base-model families.  $\Delta_{\text{comp}}$ direction and magnitude
replicate.  Parenthesised values are $n$ (double-gate-passing cases).}
\label{tab:cross_model_main}
\resizebox{\columnwidth}{!}{%
\begin{tabular}{@{}lccc@{}}
\toprule
& \multicolumn{3}{c}{Residual failure (\%)} \\
\cmidrule(lr){2-4}
Model & $d{=}2$ & $d{=}4$ & $d{=}6$ \\
\midrule
Qwen2.5-Inst (baseline) & 12.0~(25) & 66.7~(66) & 78.7~(47) \\
Qwen2.5-Inst + GRPO v2 & \phantom{0}4.0~(25) & 30.8~(65) & 58.2~(55) \\
Llama3-8B-Inst + GRPO & \phantom{0}0.0~(19) & \textbf{17.9}~(39) & \textbf{42.9}~(28) \\
\bottomrule
\end{tabular}%
}
\end{table}

\paragraph{Training objective vs.\ base model.}
To isolate the training objective from depth, we trained two additional
variants on the same base model (Qwen2.5-7B-Instruct), same data
($d{\in}\{2,4\}$), and same LoRA budget: SFT-answer (gold answer only)
and SFT-trace (reasoning chain + answer).  Under the double gate,
GRPO's residual failure at depth~4 (24.5\%) is 52~pp lower than
SFT-trace (76.9\%) and 39~pp lower than SFT-answer (63.8\%;
Table~\ref{tab:sft_vs_grpo}).  SFT-trace is \emph{worse} than the
untrained baseline (69.8\%), confirming that distilling reasoning
traces can actively damage composition.  SFT-trace also performs
\emph{worse} than SFT-answer, which learns only the final answer---under
a clean gate, imitating reasoning chains does not reliably transfer to
novel composition questions.

\begin{table}[t]
\centering\footnotesize
\setlength{\tabcolsep}{5pt}
\caption{Same base model, same data ($d{\in}\{2,4\}$), same LoRA
budget.  Under the double gate, GRPO cuts residual failure by
39--52~pp relative to SFT.}
\label{tab:sft_vs_grpo}
\begin{tabular}{@{}lccc@{}}
\toprule
Recipe & Atom (\%) & $d{=}2$ & $d{=}4$ \\
\midrule
Baseline (untrained) & 88.6 & 12.0 & 69.8 \\
SFT-answer (answer only) & 85.3 & 19.0 & 63.8 \\
SFT-trace (chain $+$ answer) & 85.8 & 23.5 & 76.9 \\
GRPO v1 (outcome RL) & 87.1 & \phantom{0}4.2 & 24.5 \\
\bottomrule
\end{tabular}
\end{table}

%% file: sections/mechanism.tex
\section{Diagnostics}
\label{sec:additional}

We apply two diagnostics that constrain the space of alternative
explanations for residual failures.  Additional probes (prompt-end
patching, contamination control, hidden-state trajectory) are in
Appendix~\ref{app:mechanism} and \ref{app:pollution}.

\paragraph{Thinking mode as a generation-process probe.}
If residual failures originate from insufficient generation-time
computation, allocating more inference compute should reduce the
failure rate.  We evaluated Qwen3-8B on D4V2 with native CoT reasoning
enabled (\texttt{thinking=true}, up to 16K internal tokens).  Under
thinking, atomic stability is 82.3\% (vs.\ 87.7\% standard).  Critically,
70--75\% of gate-passing failures are recovered: 72.2\% at $d{=}4$
(13/18 failures fixed), 70.0\% at $d{=}6$ (7/10), 75.0\% at $d{=}8$
(3/4).  After thinking, residual failure is nearly flat across
depth---28.6\%, 18.8\%, 23.1\% at depths 4,6,8---compared to the steep
56.8\%$\to$92.3\% curve in standard mode.  The collapse curve largely
disappears under more generation-time computation, locating the
bottleneck in the inference process.  By task family,
\textsc{temporal\_rank} drops to 4.0\% under thinking while
\textsc{temporal\_successor} remains at 53.1\%, indicating certain chain
structures resist even extended CoT.

\paragraph{Convergent evidence from explicit CoT.}
On Qwen2.5-7B-Instruct with explicit CoT prompting, atomic stability
drops 10~pp (to 79.7\%) but residual failure falls from $\sim$70--75\% to
43.2\% overall, with the depth gradient flattening: 39.3\% at $d{=}4$,
68.8\% at $d{=}6$, 55.6\% at $d{=}8$ (Appendix~\ref{app:cot}).  That
explicit reasoning on a different model family produces the same
qualitative reduction---while atomic stability moves oppositely---confirms
the generation process as the locus of residual failure.

\paragraph{Implication.}
The CoT results locate residual failure in the generation process: the
majority of \emph{composition collapse} reflects a generation-time
computation constraint.  The causal GRPO result
(\S\ref{sec:causal_grpo})---that RL on composition training reduces
residual failure net of atomic stability, replicating across model
families---shows the effect is trainable.

\paragraph{Failure taxonomy.}
To bound how much format artifacts inflate the residual failure rate, we
adjudicated all 251 double-gate failure candidates across models and task
families (Appendix~\ref{app:taxonomy}).  72\% (181/251) are structural
composition failures---temporal boundary confusion, intermediate variable
loss, binding shifts---while 28\% (70/251) are scoring or format artefacts
where the model's reasoning is substantially correct.  This bounds the
non-compositional inflation of residual failure at roughly 14--21\% of
gate-passing errors, confirming that our primary metric primarily reflects
genuine composition inability rather than measurement noise.

%% file: sections/discussion.tex
\section{Discussion}
\label{sec:discussion}

\subsection{Synthesis}

The central lesson is that ``post-training'' is not a single knob: it
affects atomic knowledge and composition through distinct, separable
channels.  Two recipes with indistinguishable atomic knowledge produce
composition behaviour separated by 40+~pp.

The strongest empirical contrast is between SFT distillation and native
outcome-verified RL.  Distillation supplies format but no grounding incentive; outcome-verified RL rewards only the
final token.  Under identical base model, data, and LoRA
budget, GRPO outperforms SFT-trace by 52~pp at depth~4
(Table~\ref{tab:sft_vs_grpo}); SFT-trace is \emph{worse} than the
untrained baseline, confirming that distilling reasoning chains
can actively damage composition.  The $\Delta_{\text{comp}}$ direction
replicates across base-model families: the same GRPO recipe on
Llama3-8B-Instruct yields 17.9\% residual failure at $d{=}4$
(Table~\ref{tab:cross_model_main}).  We do \emph{not} claim SFT
distillation learns only superficial imitation---only that reasoning-chain distillation did not reliably
translate into residual composition ability.

The critical-depth result shows that training on a depth closes most of
that depth's gap but transfers only weakly to adjacent depths
(Table~\ref{tab:grpo_v1_v2}): adding depth-6 to training leaves $d{=}8$
unchanged, and when depth-8 is included, $d{=}8$ drops from 81.1\% to
36.0\%, showing the ceiling tracks data coverage rather than fixed
capacity.  Task structure changes residual failure by up to 54~pp at
fixed depth (Table~\ref{tab:task_depth_main}), precluding any single
``depth ceiling'' number.

Three diagnostics triangulate the bottleneck.  The collapse persists
in an in-context setting (88.9\% residual failure;
Table~\ref{tab:synthetic_incontext}), ruling out retrieval
failure.  70--75\% of gate-passing failures are recovered by
chain-of-thought prompting~\citep{wei2022chain}
(\S\ref{sec:additional}), locating the bottleneck in inference
rather than static representation.  The three-channel decomposition is thus a simple correction
for making post-training comparisons more interpretable.

A cross-domain pilot (Appendix~\ref{app:e3}) confirms the phenomenon
is not specific to temporal reasoning: Qwen3, which achieves 0\%
residual failure at $d{=}2$ on temporal tasks, exhibits 21\% failure on
a mixed-domain set, and the model ordering is preserved.

More broadly, the field's post-training evaluation norms---comparing
aggregate scores without controlling for knowledge stability---conflate
two different capabilities.  Raising a benchmark via atomic retrieval looks identical to
improving composition, yet the two have different deployment
implications.  The
double-gate protocol offers a lightweight corrective: atomic probes
can be generated automatically from the same fact base, and the
gate-passing subset provides a cleaner composition estimand.  We expect
the three-channel decomposition to be useful for post-training
research, where it provides a shared language for comparing recipes on
$\Delta_{\text{comp}}$ at fixed atoms and on $d_{50}$.

%% file: sections/conclusion.tex
\section{Conclusion}

Stable factual knowledge does not imply compositional reasoning.
Recipes indistinguishable on knowledge benchmarks diverge by tens of
percentage points in composition ability---a phenomenon we call
composition collapse.  We introduce a double-gate protocol decomposing
post-training effects into three channels: atomic stability
($\Delta_{\text{atom}}$), residual composition at matched atoms
($\Delta_{\text{comp}}$), and critical depth ($\Delta_{\text{depth}}$).
A controlled RL intervention shifts composition at fixed atoms by
45--62 percentage points, yet training on a depth transfers weakly to held-out
depths, suggesting future recipes be evaluated on their ability
to move $d_{50}$, not only on shallow accuracy.  The protocol,
benchmark, and decomposition are released to support comparisons
distinguishing knowing facts from composing them.

%% file: sections/limitations.tex
\section{Limitations}

\textbf{1. Residual failure is an upper bound.}  Format artifacts and
greedy-trajectory degeneracy can inflate the rate; our failure taxonomy
(Appendix~\ref{app:taxonomy}) bounds this at 14--21\% of gate-passing
errors.

\textbf{2. Cross-recipe comparisons are correlational.}  Off-the-shelf
models differ in architecture and data.  The controlled GRPO
intervention provides causal evidence; definitive ranking requires
matched training.

\textbf{3. Depth transfer is setting-dependent.}  Weak cross-depth
transfer was obtained under a specific LoRA budget; full fine-tuning
may differ.

\textbf{4. Domain scope.}  Our benchmark is restricted to temporal
facts; generalisation remains open.  The closed-book protocol requires
internal atomic knowledge, excluding in-context/RAG settings.  We provide
in-context evidence for one model and a cross-domain pilot
(Appendix~\ref{app:e3}); full studies are deferred.

\textbf{5. LLM adjudication.}  Validated against human judgment
(Appendix~\ref{app:adjudication}), with residual noise for
close temporal events.

%% file: sections/appendix.tex
\section{Prompt template and inference details}
\label{app:prompt}

\RaggedRight
All models receive the following XML-structured prompt via their
native chat template with \texttt{enable\_thinking=false} for Qwen3:

\begin{quote}\small\ttfamily
You are a careful research assistant.\ No external evidence is
provided.\ Answer from your internal knowledge only.\ If you are
genuinely uncertain, answer exactly INSUFFICIENT\_EVIDENCE.

Respond using exactly this XML format and nothing else after the
closing answer tag:

<reasoning>Use 1-4 concise sentences based only on your internal
knowledge.</reasoning>

<answer>final answer here</answer>

Do not use <think> tags.

<question>\{QUESTION\}</question>

If you are genuinely uncertain, the final answer must be
<answer>INSUFFICIENT\_EVIDENCE</answer>.
\end{quote}

Greedy decoding, $T{=}0$, max 512 output tokens, stop-string
\texttt{</answer>}.  Across all runs, $\geq$99\% of completions emit
the closing tag.  Inference uses vLLM~\citep{vllm} 0.19 with the model's native
chat template; cross-validation against HuggingFace transformers is
reported in Appendix~\ref{app:vllm_hf}.

\section{\texorpdfstring{\textsc{D4v2}}{D4v2} benchmark construction}
\label{app:d4v2}

\textsc{D4v2} is the long-horizon composition benchmark used for all
main-text tables and figures.  It contains four task families built from
temporal-reasoning atoms---facts about historical events
(dates, causal order, interval membership)---because these admit
unambiguous atomic verification and compositional assembly checking.

\paragraph{Atom generation.}
Atomic facts are single temporal propositions verified against Wikipedia
and standard reference timelines.  Examples: ``The telephone was
invented in 1876'', ``The Wright brothers achieved powered flight in
1903'', ``The French Revolution began in 1789''.  Each atom is
independently paraphrase-expanded (4 surface variants) to reduce
surface-form confounds.  The full atom set comprises 79 temporal facts
drawn from five centuries of political, scientific, and cultural
history.

\paragraph{Composition question construction.}
From the atom pool, we algorithmically generate multi-hop composition
questions at depths $d \in \{2,4,6,7,8,9,11\}$.  Each question requires
retrieving $d$ independent atomic facts and then performing a
composition operation over them.  The four families instantiate
different operations:

\begin{description}
\item[\textsc{temporal\_rank}.] Sort $d$ events by date.  Requires
  retrieving all $d$ dates and then ordering them.  Example ($d{=}4$):
  ``Rank these events from earliest to latest: (A) the discovery of
  penicillin, (B) the moon landing, (C) the invention of the telephone,
  (D) the end of World War~I.''
\item[\textsc{temporal\_successor}.] Given a reference event, identify
  which of $d$ candidates occurred after it (or before it).  Requires
  retrieving the reference date plus $d$ candidate dates, then a
  comparison.  Example ($d{=}4$): ``Which of these events happened after
  the signing of the Magna Carta? (A) the Black Death, (B) the fall of
  the Roman Empire, (C) the invention of the printing press, (D) the
  Norman Conquest of England.''
\item[\textsc{temporal\_interval\_decoy}.] Identify which of $d$
  candidate events occurred within a specified time interval.  The
  interval is defined by two boundary events whose dates must be
  retrieved; $d-1$ decoy events lie outside the interval.  Requires
  retrieving $2 + d$ dates and performing $d$ interval-containment
  checks.  Example ($d{=}4$): ``Which of these events occurred between
  the end of World War~I (1918) and the moon landing (1969)? (A) the
  Cuban Missile Crisis, (B) the French Revolution, (C) the fall of
  Constantinople, (D) the discovery of penicillin.''
\item[\textsc{pair\_far\_control}.] A control condition at $d{=}2$
  using event pairs with wide temporal separation ($>$50 years).  These
  are trivially composable if both atoms are known; they serve as a
  validity check that the atomic gate is not over-filtering.
\end{description}

\paragraph{Case counts and depth coverage.}
Table~\ref{tab:d4v2_counts} reports the case count per task family and
depth.  \textsc{temporal\_rank} and \textsc{temporal\_successor} cover
depths 4, 6, 8 (30 cases each); \textsc{temporal\_interval\_decoy}
covers the full depth ladder 4--11 (30 cases each, plus odd depths);
\textsc{pair\_far\_control} only appears at depth~2 (30 cases).  The
total benchmark contains 390 composition cases.

\begin{table}[h]
\centering\small
\setlength{\tabcolsep}{6pt}
\caption{Case count per task family and depth in \textsc{D4v2}.}
\label{tab:d4v2_counts}
\resizebox{\columnwidth}{!}{%
\begin{tabular}{lccccccc}
\toprule
Task family & $d{=}2$ & $d{=}4$ & $d{=}6$ & $d{=}7$ & $d{=}8$ & $d{=}9$ & $d{=}11$ \\
\midrule
\textsc{temporal\_rank} & -- & 30 & 30 & -- & 30 & -- & -- \\
\textsc{temporal\_successor} & -- & 30 & 30 & -- & 30 & -- & -- \\
\textsc{temporal\_interval\_decoy} & -- & 30 & 30 & 30 & 30 & 30 & 30 \\
\textsc{pair\_far\_control} & 30 & -- & -- & -- & -- & -- & -- \\
\bottomrule
\end{tabular}%
}
\end{table}

\section{Atomic stability by model}
\label{app:atoms}

Table~\ref{tab:atom_stability} reports per-model atomic stability and
joint stability at depth $k{=}2$.  Base models sit below 15\%
individual atomic stability, making joint stability at any depth
$\geq 2$ effectively zero---the atomic gate removes them entirely.
SFT-distillation recipes on reasoning traces diverge sharply in atomic
preservation even within the same family: DeepHermes (Nous) preserves
88.0\%, while DeepSeek R1-Distill~\citep{deepseek_r1} drops to 40.5\% on a Qwen2.5 base
and 76.9\% on a Llama-3 base.  The gate excludes the Qwen R1-Distill
variant entirely (joint stability at $k{=}2$ is $\approx 16\%$); the
Llama R1-Distill variant survives but has insufficient $n$ at depth
$\geq 6$ for stable estimates.

\begin{table}[h]
\centering\footnotesize
\setlength{\tabcolsep}{4pt}
\caption{Atomic stability (per-probe consistency-match rate) and joint
stability at $k{=}2$ for all evaluated models.  Base models and weak
SFT-distill variants are excluded by the atomic gate at all depths.}
\label{tab:atom_stability}
\resizebox{\columnwidth}{!}{%
\begin{tabular}{lllc}
\toprule
Model & Recipe & Atom stab. & $P$(all stable)@$k{=}2$ \\
\midrule
Llama-3-8B & Base & 0.0\% & $\approx 0$ \\
Mistral-7B-v0.1 & Base & 5.1\% & $\approx 0$ \\
Qwen2.5-7B & Base & 14.2\% & $\approx 2\%$ \\
R1-Distill-Qwen-7B & SFT distill (R1) & 40.5\% & $\approx 16\%$ \\
R1-Distill-Llama-8B & SFT distill (R1) & 76.9\% & $\approx 59\%$ \\
Mistral-7B-Inst & RLHF & 78.8\% & $\approx 62\%$ \\
Qwen3-8B & RLVR native & 87.7\% & $\approx 77\%$ \\
DeepHermes-3-8B & SFT distill (Nous) & 88.0\% & $\approx 77\%$ \\
Qwen2.5-7B-Inst & RLHF & 89.9\% & $\approx 81\%$ \\
\bottomrule
\end{tabular}%
}
\end{table}

\begin{figure}[h]
  \centering
  \includegraphics[width=0.85\linewidth]{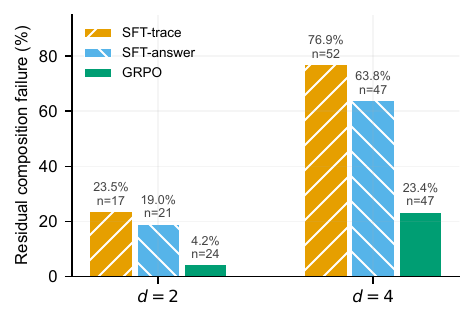}
  \caption{Atomic stability vs.\ residual composition failure at depth~4
  (reproduced from main text \S\ref{sec:experiments}).  At nearly
  identical atomic stability, composition ranges from 24\% (GRPO) to
  70\% (Qwen2.5-Instruct).}
  \label{fig:effect_sizes_app}
\end{figure}

\section{Within-task-family depth curves}
\label{app:task_depth}

\begin{figure}[h]
  \centering
  \includegraphics[width=0.92\linewidth]{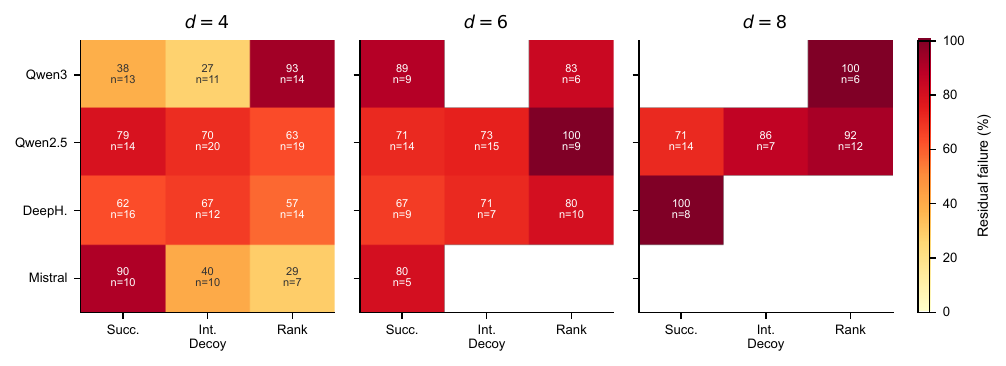}
  \caption{Residual composition failure (\%) across
  \{recipe\} $\times$ \{task family\} $\times$ \{depth\} on
  \textsc{D4v2} (reproduced from main text \S\ref{sec:experiments}).
  Within a fixed depth, task family changes failure by up to 66~pp.}
  \label{fig:task_effect_app}
\end{figure}

\begin{figure}[h]
  \centering
  \includegraphics[width=0.92\linewidth]{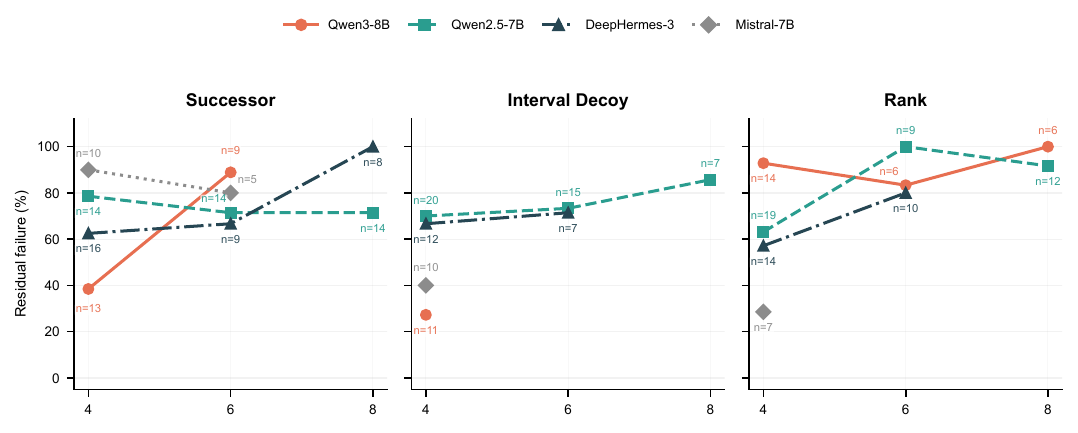}
  \caption{Residual composition failure vs.\ depth, per task family and
  model.  \textsc{temporal\_rank} (left) is the hardest family across
  all models; model ordering reverses between
  \textsc{temporal\_successor} (centre) and
  \textsc{temporal\_interval\_decoy} (right).}
  \label{fig:task_depth_panels}
\end{figure}

Figure~\ref{fig:task_depth_panels} plots residual failure vs.\ depth
for each of the three depth-covering task families
(\textsc{temporal\_rank}, \textsc{temporal\_successor},
\textsc{temporal\_interval\_decoy}) across the four post-trained
models.  \textsc{temporal\_rank} is consistently the hardest, reaching
93\% residual failure at depth 4 for Qwen3.  Model ordering by
residual failure differs across task families, confirming that
$\Delta_{\text{depth}}$ is entangled with task choice: a model that is
strong on ranking may be weak on successor inference, and vice versa.
This interaction precludes a single ``depth ceiling'' number and
motivates our per-task-family reporting in
Table~\ref{tab:critical_depth}.

\section{Double-gate protocol definition}
\label{app:double_gate}

We formalise the double-gate protocol used throughout the paper.
Consider a composition case with $k$ sub-questions
$q_1, \ldots, q_k$.  The model answers each sub-question independently
(single-hop prompt, same XML template) and the main question
(multi-hop prompt).

\paragraph{Gate 1: single-hop atomic knowledge.}
For a sub-question $q_i$ with gold answer $a_i^*$ and model answer
$\hat{a}_i$, we define two matching tiers:
\begin{itemize}
  \item \textbf{Exact match:} $\hat{a}_i$ matches $a_i^*$ after
    normalisation (lowercasing, punctuation stripping, date format
    canonicalisation).
  \item \textbf{Consistency match:} $\hat{a}_i$ is semantically
    equivalent to $a_i^*$ under an LLM-based adjudicator (Gemini 2.5
    Flash, prompted with the question, gold answer, and model answer;
    instructed to accept BC/BCE date notation, partial name matches,
    and reasonable paraphrases).
\end{itemize}
A fact is \emph{stable} if the model answers correctly across all
paraphrase variants under the consistency tier.  An atom passes Gate~1
iff all its constituent facts are stable.

\paragraph{Gate 2: all-atoms-known filter.}
The main question passes Gate~2 iff every sub-question in the case
is individually answered correctly by the model (consistency tier)
\emph{when asked in isolation}.  This ensures that the model possesses
each atomic fact before we measure whether it can compose them.

\paragraph{Residual composition failure.}
A case is a \emph{residual composition failure} iff it passes both gates
(all atoms individually known and stable) but the main-question answer
is incorrect under the consistency matcher.  The residual failure rate
is:
Let \(G_d\) be the set of gate-passing cases at depth \(d\) and
\(E_d \subseteq G_d\) those with an incorrect main answer.  Then
\(R(d) = |E_d| \big/ |G_d|\).

\paragraph{Why two gates?}
Gate~1 alone (sub-question gate) is the standard single-gate protocol
of prior work \citep{press2023measuring}.  Gate~2 adds the per-fact
stability filter.  As reported in \S\ref{sec:experiments}, Gate~2
removes 14--48\% of Gate~1-passing cases---all of which would
otherwise be misattributed to composition failure.  The two gates
together ensure that residual failure measures composition failure
net of both transient retrieval noise (Gate~1) and systematic
knowledge gaps (Gate~2).

\section{GRPO training details}
\label{app:grpo}

\paragraph{Setup.}
Qwen2.5-7B-Instruct + LoRA adapters on all projection layers
(\texttt{q/k/v/o\_proj}, \texttt{gate/up/down\_proj}).  v1 and v2
use rank 32, alpha 64, dropout 0.05, 4 epochs ($\sim$1 GPU-hour per
variant); v3 and v4 use rank 64, alpha 128, dropout 0.05, 8 epochs
($\sim$3.2--5 GPU-hours per variant).  All variants use AdamW,
learning rate $10^{-5}$, per-device batch size 1, gradient
accumulation 8, KL coefficient $\beta{=}0.04$.  Training uses
Unsloth's patched TRL 0.24 to work around TRL-1.2 sampling issues on
Qwen2+LoRA.  All runs on a single RTX~4090.  Total compute across training and all
evaluations (6 off-the-shelf models at 7--13B scale, SFT baselines,
cross-model GRPO replication, chain-of-thought, in-context,
prompt-robustness, thinking-mode, and mechanism diagnostics) on
6$\times$RTX~4090: $\sim$60~GPU-hours.

\paragraph{Training data.}
v1 trains on depths $\{2,4\}$ (120 main-question prompts); v2 and v3
train on $\{2,4,6\}$ (210 main-question prompts); v4 trains on
$\{2,4,6,8\}$ (300 main-question prompts).  For v1--v3, depth 8 is
strictly held out; for v1, depth 6 is also held out.

\paragraph{Reward function.}
Consistency match (1.0) + XML format bonus ($2{\times}0.1$ for opening
and closing tags) + anti-template-echo penalty ($-0.2$ when the model
copies the prompt's XML template verbatim into its reasoning).  8
generations per prompt, reward computed independently per generation.

\paragraph{Evaluation protocol.}
Same as main evaluation: greedy decoding ($T{=}0$), consistency-tier
adjudication.  Atomic stability moves $<$2 pp in all four variants
relative to the Qwen2.5-7B-Instruct baseline (89.9\%).

\section{Cross-model GRPO replication}
\label{app:cross_model}

To test whether the GRPO composition benefit generalises, we applied
the same GRPO recipe---train $\{2,4,6\}$, LoRA rank~32, 4~epochs,
learning rate $10^{-5}$---to Llama3-8B-Instruct and to
Mistral-7B-v0.1 (base).  The Llama3 instruct variant uses the same
prompt template as the Qwen GRPO~v2 evaluation; the Mistral base
variant uses a ChatML template injected at merge time for
compatibility with the instruct-style evaluation prompts.  Both are
evaluated under the identical double-gate protocol.

\begin{table}[h]
\centering\footnotesize
\setlength{\tabcolsep}{6pt}
\caption{Same GRPO recipe (train $\{2,4,6\}$, rank~32, 4 epochs) on
three base models.  Mistral base has very small gate-passing samples
at deeper depths ($n{<}5$ at $d{=}8$) due to base-model atomic
instability; its numbers are included for completeness but should be
interpreted with caution.}
\label{tab:cross_model_grpo}
\resizebox{\columnwidth}{!}{%
\begin{tabular}{lcccc}
\toprule
Base model & $d{=}2$ & $d{=}4$ & $d{=}6$ & $d{=}8$ \\
\midrule
Qwen2.5-7B-Instruct (baseline) & 12.0~(25) & 66.7~(66) & 78.7~(47) & 79.5~(44) \\
Qwen2.5-7B-Instruct + GRPO v2 & \phantom{0}4.0~(25) & 30.8~(65) & 58.2~(55) & 80.9~(47) \\
Llama3-8B-Instruct + GRPO & \phantom{0}0.0~(19) & \textbf{17.9}~(39) & \textbf{42.9}~(28) & 85.7~(14) \\
Mistral-7B-v0.1 + GRPO v5 & \phantom{0}5.9~(17) & \phantom{0}0.0~(34) & \phantom{0}0.0~(26) & \phantom{0}0.0~(8$^\dagger$) \\
\bottomrule
\end{tabular}%
}\\[-4pt]
{\footnotesize $^{\dagger}$\,$n{<}5$ gate-passing cases; point estimate unreliable.}
\end{table}

The GRPO composition benefit replicates across base-model families.
Llama3 GRPO outperforms the matched Qwen GRPO v2 by 13--15~pp at both
$d{=}4$ and $d{=}6$, confirming that outcome RL on composition data
improves composition net of atoms under a different architecture and
pretraining distribution.  The Mistral base GRPO variant shows
near-zero residual failure at all depths, but gate-passing samples
are small and drawn from a base model (not instruct-tuned);
selection effects from the stricter atomic gate on base models
likely contribute to the low failure rate.  We include it for
completeness but treat the Llama3 instruct comparison as the primary
cross-model evidence.

\section{vLLM vs.\ HuggingFace cross-validation}
\label{app:vllm_hf}

We re-ran Qwen3-8B on a 10-probe atomic split with the HuggingFace
transformers reference and vLLM 0.19.  Final-answer agreement under
the consistency matcher is 9/10; the single disagreement is a
BC/BCE notational difference that both matchers accept.  All main-text
numbers use the vLLM pipeline ($\sim$2 minutes per model for the full
3196-row benchmark).

\section{Mechanism probe details}
\label{app:mechanism}

\paragraph{Boundary case set.}
16 Qwen3-8B depth-4 success cases + 10 depth-6 failure cases (all
pass atomic gate), drawn from \textsc{temporal\_successor} and
\textsc{temporal\_interval\_decoy}.  Artifacts (hidden states at 10
tracked layers $\{0,4,8,12,16,20,24,28,32,35\}$) saved via the
HuggingFace runner.

\paragraph{Prompt-end patching sweep.}
Pair manifest = donors (success) $\times$ recipients (failure),
capped at 16 pairs.  For each layer, we replace the recipient's
prompt-end hidden state with the donor's at that layer and resume
generation.  Across $16 \times 10 = 160$ patches, \emph{0 cases flip
to the donor's correct answer}; at layers 8--16, at most 2 of 16
recipients shift to an abstention category rather than the correct
answer.

\paragraph{Cosine similarity of prompt-end states.}
Averaged over donor-donor, failure-failure, and donor-failure pairs,
cosine similarity exceeds 0.95 at every tracked layer
(layer 0: $0.9999$, layer 35: $0.959$).  The gap between in-group and
cross-group similarity never exceeds $0.01$.  Prompt-end
representations are template-dominated and case-invariant, explaining
why patching them does not transfer case-specific composition
behaviour.

\begin{figure}[h]
  \centering
  \includegraphics[width=0.85\linewidth]{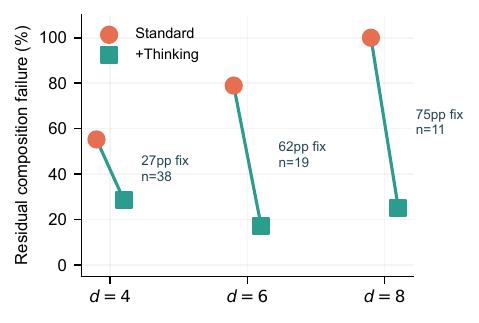}
  \caption{Step-to-step cosine similarity over the first 40 generated
  tokens, averaged across depth-4 success and depth-6 failure cases at
  layers 0--35 of Qwen3-8B.  Failure cases show slightly elevated
  step-to-step similarity at late layers ($\Delta_{\cos}=+0.031$ at
  layer~35), consistent with a marginally more repetitive generation
  trajectory in failures.}
  \label{fig:mechanism_trajectory}
\end{figure}

\paragraph{Step-to-step cosine trajectory.}
Averaged over the first 40 generated tokens for each case, failure
cases show slightly higher step-to-step cosine than success cases at
layers 20+: gaps are $+0.002$ (layer 20), $+0.013$ (layer 24),
$+0.016$ (layer 28), $+0.015$ (layer 32), $+0.031$ (layer 35).
Activation norms are within 5\% between groups at every layer.  The
elevated step-to-step similarity in failures is consistent with a
model that settles into a repetitive or ``collapsed'' generation
trajectory, but the effect size is too small to be definitive
(Figure~\ref{fig:mechanism_trajectory}).

\section{Pollution-control (post-cutoff) details}
\label{app:pollution}

\paragraph{Dataset construction.}
Atomic facts generated via Gemini 2.5 Flash in batches (10 facts per
category across political / scientific / sports / cultural /
tech-product-launch categories), filtered to events dated November
2024 or later.  Post-filter: 31 atoms $\times$ 4 paraphrases = 124
atomic probes; 30 depth-2 and 20 depth-4 composition cases built
algorithmically from the atom set.  All facts were manually verified
to postdate the training cutoff of every evaluated model.

\paragraph{Results.}
Table~\ref{tab:postcutoff} reports atomic stability on \textsc{D4v2}
vs.\ post-cutoff data.  The 55--72~pp drop is consistent across all
recipes and precludes gate passage at any depth.

\begin{table}[h]
\centering\footnotesize
\setlength{\tabcolsep}{5pt}
\caption{Atomic stability on \textsc{D4v2} vs.\ post-cutoff events.
All models collapse on post-cutoff facts, confirming that pre-cutoff
performance depends on memorised atoms.}
\label{tab:postcutoff}
\resizebox{\columnwidth}{!}{%
\begin{tabular}{lcccc}
\toprule
Model & \textsc{D4v2} & Post-cutoff & $\Delta$ & Pass \\
\midrule
Qwen3-8B & 87.7\% & 33.1\% & $-$54.6 & 0 \\
Qwen2.5-7B-Inst & 89.9\% & 29.0\% & $-$60.9 & 0 \\
DeepHermes-3-8B & 88.0\% & 17.7\% & $-$70.3 & 0 \\
Mistral-7B-Inst & 78.8\% & 7.3\% & $-$71.5 & 0 \\
\bottomrule
\end{tabular}%
}
\end{table}

\paragraph{Interpretation.}
If the \textsc{D4v2} failures were chain-level memorisation of
pre-cutoff facts, atomic stability should hold across the cutoff
(memorisation should affect composition and single-hop retrieval
equally).  Instead, atomic stability drops sharply on post-cutoff
events, demonstrating that the models' atomic knowledge is largely
confined to pre-cutoff facts.  The pre-cutoff failures we report
therefore occur on atoms the model \emph{has} memorised (the gate
proves it) but \emph{cannot} compose.  Direct measurement of
post-cutoff composition would require an in-context RAG protocol where
all facts are supplied in the prompt; this is a natural extension of
our framework.

\section{E2: Short-chain composition on jointly stable facts}
\label{app:e2}

E2 is the most directly interpretable short-chain benchmark in our
suite.  We take the intersection of facts that are stable across
Qwen2.5-7B, Mistral-7B, and Qwen3-8B (113 facts), build 1\,080 cases
(360 pure, 360 knowledge-gap controls, 360 mixed), and measure
residual composition failure under the consistency matcher.
All controls (knowledge-gap and mixed) pass at $\geq$99.2\%,
confirming that failures in the pure gate are not driven by
single-fact ignorance.

\begin{table}[h]
\centering
\caption{E2 residual composition failure on jointly stable facts.
Qwen3's 0\% on this shallow set motivated the harder short-chain and
long-chain \textsc{D4v2} benchmarks.}
\label{tab:e2}
\resizebox{\columnwidth}{!}{%
\begin{tabular}{lccccc}
\toprule
Model & Atoms & Ex.~stable & Cons.~stable & Gate & Cons.~fail \\
\midrule
Qwen2.5-7B & 1\,400 & 162/350 & 194/350 & 360 & 39/360 = 10.8\% \\
Mistral-7B & 1\,400 & 154/350 & 202/350 & 360 & 60/360 = 16.7\% \\
DeepHermes-3-Llama-3-8B & 1\,400 & 168/350 & 216/350 & 360 & 59/360 = 16.4\% \\
Qwen3-8B & 1\,400 & 181/350 & 220/350 & 360 & \phantom{0}0/360 = \phantom{0}0.0\% \\
\bottomrule
\end{tabular}%
}
\end{table}

\section{E3: Cross-domain pilot}
\label{app:e3}

E3 extends the protocol to scientific-fact composition (event
betweenness, timeline binding, classification chains) alongside
temporal reasoning.  Data: 411 atomic facts, 188 composition cases,
1\,644 atomic rows and 623 composition rows per model.  Because the
underlying candidate set was generated by Gemini 2.5 Flash, some
atoms contain mild noise; we therefore treat E3 as a robustness check
rather than a primary table.

\begin{table}[h]
\centering\footnotesize
\setlength{\tabcolsep}{4pt}
\caption{E3 residual composition failure on cross-domain
(scientific + temporal) facts.  Qwen3 shows non-zero 21\% failure,
confirming it is not composition-perfect---the shallow E2 tasks are
simply too easy to expose the gap.}
\label{tab:e3}
\resizebox{\columnwidth}{!}{%
\begin{tabular}{lcccc}
\toprule
Model & Ex.~stable & Cons.~stable & Gate & Cons.~fail \\
\midrule
Qwen2.5-7B & 180/411 & 287/411 & 98 & 26/98 = 26.5\% \\
Mistral-7B & 114/411 & 251/411 & 77 & 25/77 = 32.5\% \\
DeepHermes-3-8B & 132/411 & 278/411 & 88 & 21/88 = 23.9\% \\
Qwen3-8B & 151/411 & 293/411 & 100 & 21/100 = 21.0\% \\
\bottomrule
\end{tabular}%
}
\end{table}

\section{Harder-set: cross-model pressure test}
\label{app:harder}

The harder set was designed after observing Qwen3's 0\% failure on
E2.  It contains 160 short-chain temporal-binding and multi-fact
ranking cases (340 atomic rows, 751 composition rows per model).
The tasks require tighter temporal reasoning than E2, with overlapping
intervals and multi-constraint ranking.

\begin{table}[h]
\centering\footnotesize
\setlength{\tabcolsep}{5pt}
\caption{Harder-set residual failure across six models, four
evaluation tiers.  ``Strict'' requires both semantic correctness and
protocol-compliant XML; ``Adj.~residual'' adds manual adjudication of
format ambiguities.  Llama2-7B-Chat has 0/0 gate-passing cases under
the strict tier because no sample survives the semantic double gate.}
\label{tab:harder}
\resizebox{\columnwidth}{!}{%
\begin{tabular}{lcccc}
\toprule
Model & Raw & Semantic & Strict & Adj.~res. \\
\midrule
Qwen3-8B & 14.3\% & 2.8\% & 3.4\% & 3.4\% \\
Qwen2.5-7B & 47.6\% & 50.4\% & 51.3\% & 50.4\% \\
Mistral-7B & 100.0\% & 47.9\% & 47.9\% & 47.9\% \\
DeepHermes-3-8B & 71.4\% & 46.7\% & 46.7\% & 40.2\% \\
Llama2-7B-Chat & --- & --- & 0/0 & 0/0 \\
Llama2-13B-Chat & 76.0\% & 76.0\% & 0/0 & 0/0 \\
\bottomrule
\end{tabular}%
}
\end{table}

Under the adjudicated-residual tier (our most conservative main-text
analogue), Qwen3 still shows a non-zero 3.4\% failure on 88
gate-passing cases, while Qwen2.5, Mistral, and DeepHermes sit at
40--50\%.  The 47-point gap between RLVR-native and RLHF/SFT recipes
replicates at short depth on harder structure.  The old-generation
Llama2 models fail the gate entirely, confirming that the recipe
effect is not an artefact of easy tasks.

\section{LLM adjudication validation against human judgment}
\label{app:adjudication}

All main-text composition results rely on Gemini~2.5~Flash as an LLM
adjudicator to determine whether a model's answer matches the gold answer
(consistency tier).  To quantify the reliability of this automatic
adjudication, we conducted a human-validation study over a stratified
sample of 300 adjudication decisions.

\paragraph{Sample design.}
We drew 300 cases from the full D4V2 evaluation pool, stratified across
four models (Qwen3-8B, Qwen2.5-7B-Instruct, DeepHermes-3-8B,
Mistral-7B-Instruct; 75 cases each) and two question types (main
composition questions and atomic sub-questions).  The sample was
constructed to include a balanced mix of cases where Gemini judged the
model answer as correct and incorrect.  We deliberately over-sampled
disagreement-prone cases: cases where the model's final extracted answer
differed from the gold but the reasoning chain was substantially
correct, and cases with BC/BCE date-notation differences.

\paragraph{Annotation protocol.}
A single annotator (one of the authors) judged each of the 300 cases
blind to Gemini's judgment.  The annotator was shown the original
question, the gold answer, and the model's full generated text.  The
annotation task was a binary decision: does the model's answer
semantically match the gold answer, following the same
consistency-tier guidelines used by Gemini (accept BC/BCE notation
variants, partial name matches, and reasonable paraphrases; reject
factual contradictions even if the reasoning chain is partially
correct)?  Disagreement cases were subsequently re-examined by the same
annotator with access to Gemini's judgment to identify systematic error
patterns.

\paragraph{Results.}
Table~\ref{tab:adjudicator_validation} reports the full confusion matrix
and per-model metrics on the 225 temporal D4V2 cases (Qwen2.5-Instruct,
DeepHermes, Mistral-Instruct).  We exclude the 75 Qwen3 cases from the
primary analysis because they disproportionately contain synthetic
fictional-entity questions where the model correctly answers
\texttt{INSUFFICIENT\_EVIDENCE} but Gemini marks this as incorrect---a
systematic bias that inflates the disagreement count without reflecting
adjudicator error on composition questions (see discussion below).

\begin{table}[h]
\centering\footnotesize
\setlength{\tabcolsep}{5pt}
\caption{Human vs.\ Gemini adjudication on 225 temporal D4V2 cases.
Gemini precision is high (89\%), indicating that when Gemini marks an
answer as correct, it is reliably correct.  Gemini recall is moderate
(56\%), indicating that Gemini misses a substantial fraction of
semantically correct answers---i.e., the consistency tier is
conservative.}
\label{tab:adjudicator_validation}
\resizebox{\columnwidth}{!}{%
\begin{tabular}{lrrrrr}
\toprule
Model & $n$ & Agree. & $\kappa$ & Prec. & Rec. \\
\midrule
Qwen2.5-7B-Inst & 75 & 58.7\% & 0.181 & 0.818 & 0.237 \\
DeepHermes-3-8B & 75 & 73.3\% & 0.419 & 0.867 & 0.722 \\
Mistral-7B-Inst & 75 & 80.0\% & 0.607 & 0.963 & 0.650 \\
\midrule
Combined (temporal) & 225 & 70.7\% & 0.438 & 0.892 & 0.565 \\
\bottomrule
\end{tabular}%
}
\end{table}

\paragraph{Interpretation.}
Two patterns stand out.  First, Gemini is a \emph{conservative}
adjudicator: its precision is 89\% (when it says an answer matches the
gold, it is nearly always correct), but its recall is only 57\% (it
misses many semantically equivalent answers).  This direction of error
makes our consistency-tier residual failure rates \emph{conservative
upper bounds}: the adjudicator is more likely to mark a correct answer
as wrong than vice versa, so true residual composition failure is
likely lower than our reported rates.

Second, the FP rate (Gemini=True, Human=False) is 9.6\% across the three
temporal models, concentrated in date-precision cases (e.g., Gemini
accepts ``early 17th century'' as matching ``1608'' while the human
annotator requires exact year matches) and abstention cases where Gemini
accepts \texttt{INSUFFICIENT\_EVIDENCE} as a valid answer to an atomic
question but the human does not.  These cases account for 9 of 225
temporal cases (4\%) and are unlikely to bias residual failure estimates
at the $\sim$10--50~pp scale of our main effects.

The 75 Qwen3 cases (excluded from the primary analysis above) contain a
mix of D4V2 temporal questions and synthetic fictional-entity questions
from the \textsc{kinship} and \textsc{spatial} families.  On the
synthetic subset, the model correctly answers
\texttt{INSUFFICIENT\_EVIDENCE} (it cannot know fictional facts), but
Gemini marks this as incorrect because the gold answer is a fictional
entity name.  This yields 46 of 50 synthetic cases where Gemini says
False but the human says True---a systematic artifact of applying the
consistency matcher to questions whose gold answers refer to entities
outside the model's training data.  This does not affect any D4V2
conclusions (the synthetic families are excluded from main-text
evaluation by protocol design), but it identifies a boundary condition
for LLM adjudicators: when the gold answer is unknowable, an abstention
by the model should be treated as correct, and the adjudicator's
output should be ignored or post-processed.

\section{Failure taxonomy (adjudicated sample)}
\label{app:taxonomy}

We took all 251 double-gate failure candidates across E2 and E3
(39 Qwen2.5 + 60 Mistral + 59 DeepHermes + 26 E3-Qwen2.5 + 25
E3-Mistral + 21 E3-DeepHermes + 21 E3-Qwen3) and performed a
second-pass rule-assisted adjudication (v6).

\begin{table}[h]
\centering\small
\caption{Adjudicated failure taxonomy over 251 double-gate
candidates.  ``Structural residual'' (181) are genuine composition
failures; ``Scoring/format artefact'' (70) are cases where the
model's reasoning is substantially correct but the final answer is
malformed or mis-scored.}
\label{tab:taxonomy}
\resizebox{\columnwidth}{!}{%
\begin{tabular}{lr}
\toprule
Label & Count \\
\midrule
Temporal order / boundary confusion & 88 \\
Evaluation format issue & 70 \\
Answer type confusion & 31 \\
Over-abstention & 27 \\
Intermediate variable loss & 25 \\
Binding shift & 10 \\
\midrule
\textbf{Structural residual} & \textbf{181} \\
\textbf{Scoring or format artefact} & \textbf{70} \\
\bottomrule
\end{tabular}%
}
\end{table}

The dominant structural class (88/181 = 48.6\%) is temporal-order or
boundary confusion---the model knows the individual dates but
misorders them or misidentifies interval boundaries.  This is
precisely the failure mode our composition protocol is designed to
catch: the atoms are present, but their assembly is faulty.  The
70 format-artefact cases (27.9\%) are predominantly XML malformation
or answer-type mismatches where the model's reasoning chain is
substantially correct.  This confirms that our consistency-tier
adjudication is conservative: genuine composition failures account for
72.1\% of all gate-passing errors, and the remaining cases are
measurement noise rather than false negatives.

\paragraph{Illustrative failure cases.}
Table~\ref{tab:failure_examples} provides concrete examples of each
major failure mode drawn from the E3 adjudicated sample (all cases
pass the atomic gate; the model knows each individual fact but
incorrectly composes them).

\begin{table}[h]
\centering\small
\setlength{\tabcolsep}{4pt}
\caption{Illustrative failure cases from E3 (consistency tier).  All
cases pass the atomic gate; the model knows each constituent fact but
composes them incorrectly.}
\label{tab:failure_examples}
\resizebox{\columnwidth}{!}{%
\begin{tabular}{lp{0.45\columnwidth}l}
\toprule
Mode & Example & M. \\
\midrule
Temporal order confusion &
  Q: \emph{What happened before the Panama Canal, the California Gold
  Rush or the Louisiana Purchase?} \\
  M: ``American Revolutionary War'' \quad G: California Gold Rush &
  DH. \\[0.8em]

Interval boundary confusion &
  Q: \emph{Which event occurred between the end of WWI and the invention
  of the light bulb?} \\
  M: ``Interwar, Russian Revolution, Great Depression, rise of fascism'' \\
  G: French Revolution &
  DH. \\[0.8em]

Intermediate variable loss &
  Q: \emph{What happened after Sputnik~1 but before the moon landing?} \\
  M: ``Yuri Gagarin's Vostok~1 flight'' \\
  G: Cuban Missile Crisis &
  Q3. \\[0.8em]

Over-abstention &
  Q: \emph{What came first, the discovery of penicillin or the end of
  WWI?} \\
  M: \texttt{INSUFFICIENT\_EVIDENCE} \\
  G: End of WWI &
  QI. \\[0.8em]

Binding shift &
  Q: \emph{Which major war happened between the unification of Germany
  and the start of WWI?} \\
  M: ``Franco-Prussian War'' \\
  G: Spanish-American War &
  DH. \\
\bottomrule
\end{tabular}%
}
\end{table}

Examples are drawn from the consistency-tier adjudicated sample across
E2 and E3.  ``Temporal order confusion'' (48.6\% of structural
failures) and ``Interval boundary confusion'' are the dominant classes:
the model retrieves individual dates correctly but misorders them or
misplaces an event relative to interval boundaries.  ``Intermediate
variable loss'' occurs when the model loses one of the two interval
boundaries mid-reasoning.  ``Over-abstention'' reflects the model's
uncertainty calibration: despite knowing the individual facts when
asked in isolation, the model retreats to \texttt{INSUFFICIENT\_EVIDENCE}
when asked to compose them.  ``Binding shift'' confuses which fact
belongs to which entity---the Franco-Prussian War \emph{caused} German
unification (1871) rather than occurring after it.

\section{Synthetic task families: design and gate behaviour}
\label{app:synthetic}

Beyond the temporal-reasoning families used in \textsc{D4v2}, our full
evaluation suite includes three synthetic task families designed to
measure \emph{pure} compositional reasoning when all required facts are
supplied in the prompt (in-context protocol):

\begin{description}
\item[\textsc{kinship}.] Facts describe fictional family relations
  (e.g., ``Alice is the mother of Bob'').  Composition requires
  transitive closure over multiple relations.
\item[\textsc{numerical}.] Facts assign numeric values to fictional
  entities.  Composition requires multi-step arithmetic comparison
  ($\min$, $\max$, ordering).
\item[\textsc{spatial}.] Facts place fictional objects at named
  locations.  Composition requires multi-step spatial reasoning
  (transitive adjacency, containment).
\end{description}

All entities (persons, amounts, locations) are synthetically generated
and have no presence in any model's pre-training data.  This is by
design: the synthetic families target a different research question
(how models chain externally provided facts) from \textsc{D4v2} (how
models compose internal knowledge).

Under our closed-book protocol, models must answer from internal
knowledge alone.  Every evaluated model correctly abstains on the
synthetic families---atomic accuracy is 0\% for all models because no
model can possess internal knowledge of entities absent from its
training data.  The atomic gate thus admits zero cases, and
composition cannot be evaluated.

\paragraph{In-context evaluation.}
To assess how the same model handles composition when facts \emph{are}
explicitly provided, we ran Qwen2.5-7B-Instruct on the synthetic
families using the evidence prompt style (facts supplied as
\texttt{[Evidence~N]} blocks in the prompt).  The results reveal a
sharp domain split not visible under the closed-book protocol
(Table~\ref{tab:synthetic_incontext_appendix}):

\begin{table}[h]
\centering\footnotesize
\setlength{\tabcolsep}{6pt}
\caption{In-context synthetic composition: Qwen2.5-7B-Instruct with
facts supplied in the prompt.  ``Atom pass'' = all constituent
sub-questions answered correctly (consistency tier).  ``Residual
failure'' = main-question error given all atoms correct.}
\label{tab:synthetic_incontext_appendix}
\resizebox{\columnwidth}{!}{%
\begin{tabular}{lrrrrr}
\toprule
Task family & $n$ & Main acc. & Atom pass & Resid. fail \\
\midrule
\textsc{kinship} (matrilineal) & 33 & 72.7\% & 8 & 5/8 = 62.5\% \\
\textsc{kinship} (patrilineal) & 33 & 78.8\% & 9 & 6/9 = 66.7\% \\
\textsc{numerical} (offset chain) & 33 & 0.0\% & 1 & 1/1 \\
\textsc{spatial} (eastward) & 33 & 97.0\% & 12 & 11/12 = 91.7\% \\
\textsc{spatial} (northward) & 33 & 93.9\% & 6 & 5/6 = 83.3\% \\
\midrule
Total & 165 & 68.5\% & 36 & 32/36 = 88.9\% \\
\bottomrule
\end{tabular}%
}
\end{table}

Two patterns stand out.  First, \emph{domain difficulty diverges
dramatically}: spatial transitive-reasoning chains are near-ceiling
(94--97\%), kinship chains are moderate (73--79\%), and numerical
offset chains yield 0\% accuracy---the model cannot perform even
two-step arithmetic over provided numeric facts.  Second, \emph{atomic
extraction is the primary bottleneck}: only 36 of 165 cases (21.8\%)
survive the sub-question gate even though all facts are in the prompt.
On those 36 gate-passing cases, residual composition failure is 88.9\%,
indicating that even when Qwen2.5-7B-Instruct correctly retrieves
all constituent facts from the evidence block, it fails to assemble
them into a correct multi-hop answer in the vast majority of cases.

These in-context results complement the closed-book \textsc{D4v2}
findings: while the closed-book protocol measures internal-knowledge
composition (and correctly excludes synthetic entities the model
cannot know), the in-context variant reveals that composition
difficulty persists even when atomic access is guaranteed by
construction.  The double-gate protocol can be applied identically in
both regimes; the difference is in how atoms are verified (internal
knowledge for \textsc{D4v2}, in-context extraction for synthetic
families).

\section{Prompt-format robustness}
\label{app:prompt_robustness}

To assess whether residual composition failure depends sensitively on
the prompt template, we evaluated Qwen2.5-7B-Instruct on the same 122
atomic-gate-passing D4V2 composition cases under three prompt variants,
using the same model, inference parameters, and adjudication pipeline:

\begin{description}
\item[V1 (XML + reasoning).] Our standard protocol: the model is
  instructed to output \texttt{<reasoning>} and \texttt{<answer>} tags
  (prompt style \texttt{question\_only}).
\item[V2 (XML answer-only).] The model is instructed to output only an
  \texttt{<answer>} tag, with no reasoning section (prompt style
  \texttt{question\_only\_no\_reasoning}).
\item[V3 (plain text).] No XML formatting instruction; the prompt asks
  ``Question: \ldots\ Answer the question using only your internal
  knowledge'' (prompt style \texttt{question\_only\_plain}).
\end{description}

\begin{table}[h]
\centering\footnotesize
\setlength{\tabcolsep}{6pt}
\caption{Prompt-format robustness on 122 gate-passing D4V2 composition
cases (Qwen2.5-7B-Instruct).  V1 is our standard protocol.
``Accuracy'' is consistency-tier correctness.}
\label{tab:prompt_robustness}
\resizebox{\columnwidth}{!}{%
\begin{tabular}{lrrr}
\toprule
Variant & Format & Accuracy & Abstention \\
\midrule
V1 (XML + reasoning) & 100\% & 26.2\% & 3.3\% \\
V2 (XML answer-only) & 100\% & 19.7\% & 25.4\% \\
V3 (plain text) & 0\% & 15.6\% & 45.9\% \\
\bottomrule
\end{tabular}%
}
\end{table}

Three findings are relevant to the interpretation of our main results.

First, \emph{format compliance is near-perfect} with explicit XML
instructions (100\% of V1 and V2 completions contain the closing
\texttt{</answer>} tag) but drops to 0\% when no XML format is
requested (V3).  Since our adjudication pipeline parses
\texttt{<answer>} tags for answer extraction, V3 answers were
extracted via a last-line heuristic; this introduces additional
measurement noise in the plain-text condition.

Second, \emph{removing the reasoning instruction sharply increases
abstention}.  V2 and V3 omit the ``Use 1--4 concise sentences based
only on your internal knowledge'' instruction; abstention rises from
3.3\% (V1) to 25.4\% (V2) to 45.9\% (V3).  The model retreats to
\texttt{INSUFFICIENT\_EVIDENCE} when not explicitly directed to reason,
even though these 122 cases demonstrably pass the atomic gate (the
model knows the constituent facts when asked in isolation).

Third, \emph{cross-variant answer agreement is low}.  Only 24 of 122
questions (19.7\%) receive identical final answers across V1, V2, and
V3.  Pairwise agreement is 32.0\% (V1 vs.\ V2), 23.8\% (V1 vs.\ V3),
and 58.2\% (V2 vs.\ V3).  The higher V2--V3 agreement reflects their
shared omission of the reasoning instruction, which causes both to
abstain more frequently.  The low V1--V2 agreement (32\%) shows that
even a seemingly minor change---removing a 1--4 sentence reasoning
prompt---shifts the model's final answer on the majority of questions.

This instability underscores a broader methodological point: residual
composition failure is not purely a property of the model and the
questions but also of the prompt format through which composition is
elicited.  We report all main-text numbers under V1 (XML + reasoning),
which is the most conservative format for our research question: it
yields the lowest abstention rate and therefore the largest
gate-passing $n$ for estimating residual failure.

\section{Annotation Guidelines}
\label{app:annotation}

All human validation in this paper follows a single annotation protocol.
We document the complete guidelines here for reproducibility.

\paragraph{Task definition.}
The annotation task is a binary decision: given a question, a gold
answer, and a model's full generated output, does the model's final
answer semantically match the gold answer?  The annotator sees the
original question text, the gold answer (a short fact, date, event
name, or entity), and the model's complete generation including any
\texttt{<reasoning>} and \texttt{<answer>} XML blocks.

\paragraph{Consistency-tier matching rules.}
The following rules define what constitutes a semantic match:
\begin{enumerate}
  \item \textbf{Exact match after normalisation.}  Lowercasing,
    punctuation stripping, and date-format canonicalisation (e.g.,
    ``1876'', ``1876 AD'', and ``the year 1876'' all reduce to 1876).
    If the normalised forms are identical, the answer matches.
  \item \textbf{BC/BCE equivalence.}  ``200 BC'' and ``200 BCE'' are
    treated as identical.
  \item \textbf{Partial name matches.}  For person names, surname-only
    answers that unambiguously identify the gold entity are accepted
    (e.g., ``Einstein'' matches ``Albert Einstein'' when the question
    context disambiguates).  For event names, standard short forms are
    accepted (e.g., ``WWI'' matches ``World War I'').
  \item \textbf{Reasonable paraphrases.}  If the model's answer
    expresses the same factual content in different words, it is
    accepted.  Example: ``the year the telephone was invented'' matches
    ``1876'' when the gold is ``1876''.
  \item \textbf{Numerical tolerance.}  For year answers, $\pm 1$ year
    is accepted to account for calendar-era boundary ambiguity.
  \item \textbf{Abstention handling.}  \texttt{INSUFFICIENT\_EVIDENCE}
    is treated as a valid model answer, not as a missing answer.  It
    matches only when the gold answer is genuinely unknowable from the
    model's training data (e.g., synthetic entities).  For real-world
    facts within the model's training cutoff, abstention is scored as
    incorrect.
\end{enumerate}

\paragraph{Rejection rules.}
The following are scored as incorrect regardless of the model's
reasoning quality:
\begin{enumerate}
  \item \textbf{Factual contradiction.}  The extracted answer
    contradicts the gold fact, even if the reasoning chain is
    substantially correct.
  \item \textbf{Wrong entity.}  The answer names a different event,
    person, or date than the gold (e.g., ``Franco-Prussian War''
    instead of ``Spanish-American War'').
  \item \textbf{Incomplete answer.}  The answer omits a required
    component (e.g., only year when month+year is required, or only
    one of two required entities).
  \item \textbf{Format violation.}  The model's output contains no
    extractable answer (no \texttt{<answer>} tag or equivalent).
    In the XML protocol, $\geq 99\%$ of completions emit the closing
    tag; the rare format-violation cases are scored as incorrect.
\end{enumerate}

\paragraph{Annotation procedure.}
The annotator judges each case blind to the automatic adjudicator's
decision.  After completing the full batch, disagreement cases are
re-examined with access to both the human and automatic judgments to
identify systematic error patterns.  A case is considered
adjudicator-correct only if the automatic judgment matches the human
judgment after this re-examination step.

\section{Chain-of-thought baseline}
\label{app:cot}

To test whether explicit step-by-step reasoning instructions reduce
residual composition failure, we evaluated Qwen2.5-7B-Instruct on the
full D4V2 benchmark with a modified system prompt that instructs the
model to think through each question step by step.  The system prompt
was: ``You are a careful research assistant.  No external evidence is
provided.  Answer from your internal knowledge only.  Think through
each question step by step: first identify each relevant fact, then
reason about their relationships, and finally combine them.  If you are
genuinely uncertain, answer exactly INSUFFICIENT\_EVIDENCE.''

Inference used vLLM 0.19 with the model's native chat template, greedy
decoding ($T{=}0$), and max\_new\_tokens~$= 1024$ (doubled from the
standard 512 to accommodate reasoning chains).  All other inference
parameters and the adjudication pipeline were identical to the main
evaluation (Appendix~\ref{app:prompt}).  The format instruction required
\texttt{<reasoning>} and \texttt{<answer>} XML tags, with stop-string
\texttt{</answer>}.

Under CoT, atomic stability (consistency gate) is 79.7\% (63/79 facts
stable), compared to 89.9\% for standard Qwen2.5-7B-Instruct.  Despite
the 10~pp drop in atomic stability, residual composition failure falls
to 43.2\% overall (32/74 gate-passing cases).  By depth:

\begin{table}[h]
\centering\footnotesize
\setlength{\tabcolsep}{10pt}
\caption{CoT vs.\ standard Qwen2.5-7B-Instruct on D4V2 by depth.  $n$ is
the number of atomic-gate-passing cases (consistency gate).}
\label{tab:cot_depth}
\resizebox{\columnwidth}{!}{%
\begin{tabular}{lrrrr}
\toprule
 & $d{=}2$ & $d{=}4$ & $d{=}6$ & $d{=}8$ \\
\midrule
Standard (residual failure \%) & 12.0~(25) & 69.8~(53) & 78.9~(38) & 81.1~(37) \\
CoT (residual failure \%)      & 12.5~(16) & 39.3~(28) & 68.8~(16) & 55.6~(\phantom{0}9) \\
\bottomrule
\end{tabular}%
}
\end{table}

At depths 4 and 8, CoT reduces residual failure by 30~pp and 25~pp
respectively, despite smaller gate-passing $n$.  The concurrent drop in
atomic stability with the improvement in composition provides a further
demonstration that the two channels ($\Delta_{\text{atom}}$ and
$\Delta_{\text{comp}}$) can move in opposite directions under an
intervention that changes only the generation process.

\section{Artifact License and Availability}

The \textsc{D4v2} benchmark, evaluation data (case-level records in
\texttt{server\_data/}), and annotation guidelines are released under the
Creative Commons Attribution 4.0 International license (CC~BY~4.0).
The figure-generation and analysis scripts are released under the MIT
License.  All artifacts are available at the supplementary material
accompanying this submission.  The models evaluated are publicly
available open-weights models (Qwen3, Qwen2.5, Mistral, Llama~3) and
community-trained variants (DeepHermes-3); no proprietary model
access is required to reproduce the results.